\documentclass{ieeeaccess}
\usepackage{cite}
\usepackage{amsmath,amssymb,amsfonts}
\usepackage{graphicx}
\usepackage{textcomp}

\usepackage{algorithm}
\usepackage{listings}
\usepackage{lipsum}
\usepackage{multirow}
\usepackage{lineno,hyperref}
\usepackage{algpseudocode}
\usepackage{booktabs}
\usepackage{enumitem}
\usepackage{amsmath}
 
\usepackage{soul} 

\usepackage{caption}
\DeclareCaptionFont{ieeeblue}{\color{accessblue}}
\DeclareCaptionLabelFormat{myformat}{\raggedright\figcapfont{\textbf{#1}\textbf{#2}}}
\renewcommand{\lstlistingname}{LISTING}
\def \hcaption {7pt}

\def\BibTeX{{\rm B\kern-.05em{\sc i\kern-.025em b}\kern-.08em
 T\kern-.1667em\lower.7ex\hbox{E}\kern-.125emX}}
\begin{document}

\renewcommand{\lstlistingname}{ALGORITHM}

\history{Received November 13, 2020, accepted December 24, 2020, date of publication December 30, 2020, date of current version January 6, 2021.}
\doi{10.1109/ACCESS.2020.3048194}

\title{Self-Corrective Sensor Fusion for Drone Positioning in Indoor Facilities}
\author{Francisco Javier González-Castaño\authorrefmark{1}, Felipe Gil-Castiñeira\authorrefmark{1}, David Rodríguez-Pereira\authorrefmark{1}, José Ángel Regueiro-Janeiro\authorrefmark{1}, Silvia García-Méndez\authorrefmark{1}, and David Candal-Ventureira\authorrefmark{1}}
\address[1]{Information Technologies Group, atlanTTic, University of Vigo, Spain}
\tfootnote{This work was partially supported by Ministerio de Economía, Industria y Competitividad, Spain, under grant TEC2016-76465-C2-2-R and Xunta de Galicia under grant GRC2018/053.}

\markboth
{Francisco Javier González-Castaño \headeretal: Self-Corrective Sensor Fusion for Drone Positioning in Indoor Facilities}
{Francisco Javier González-Castaño \headeretal: Self-Corrective Sensor Fusion for Drone Positioning in Indoor Facilities}

\corresp{Corresponding author: Francisco Javier González-Castaño (e-mail: javier@det.uvigo.es).}

\begin{abstract}
Drones may be more advantageous than fixed cameras for quality control applications in industrial facilities, since they can be redeployed dynamically and adjusted to production planning. The practical scenario that has motivated this paper, image acquisition with drones in a car manufacturing plant, requires drone positioning accuracy in the order of 5 cm. During repetitive manufacturing processes, it is assumed that quality control imaging drones will follow highly deterministic periodic paths, stop at predefined points to take images and send them to image recognition servers. Therefore, by relying on prior knowledge about production chain schedules, it is possible to optimize the positioning technologies for the drones to stay at all times within the boundaries of their flight plans, which will be composed of stopping points and the paths in between. This involves mitigating issues such as temporary blocking of line-of-sight between the drone and any existing radio beacons; sensor data noise; and the loss of visual references. We present a self-corrective solution for this purpose. It corrects visual odometer readings based on filtered and clustered Ultra-Wide Band (UWB) data, as an alternative to direct Kalman fusion. The approach combines the advantages of these technologies when at least one of them works properly at any measurement spot. It has three method components: independent Kalman filtering, data association by means of stream clustering and mutual correction of sensor readings based on the generation of cumulative correction vectors. The approach is inspired by the observation that UWB positioning works reasonably well at static spots whereas visual odometer measurements reflect straight displacements correctly but can underestimate their length. Our experimental results demonstrate the advantages of the approach in the application scenario over Kalman fusion, in terms of stopping point detection and trajectory estimation error.
\end{abstract}


\titlepgskip=-15pt

\maketitle

\section{Introduction}
\label{intro}
In general, three main applications have been proposed for drones in industrial scenarios \cite{hoffmann2017inventory,Macoir2019}: surveillance (both for security and safety), just-in-time part delivery (in which the drones carry the parts) and inventory control (where drones scan the identifiers of items beyond manual reach). In these scenarios, drone payloads will be typically light, consisting in imaging cameras (for surveillance or barcode scanning), data communication units and limited onboard processing power. Delivery drones are an exception\footnote{{\tt https://www.digitaltrends.com/cars/
zf-drone-delivery-factories-germany/}, November 2020.}. These, and in general all transport drones \cite{Kellermann2020}, need powerful battery packs, suitable motors and actuators to carry significant weights. They have been proposed for logistics \cite{Aurambout2019} and healthcare \cite{Amukele2015}. Imaging applications in open spaces are the most feasible of these scenarios. Cheap drones can easily support them and drone cameras may compensate even for coarse positioning errors. So far, in the most realistic industrial applications of drones, piloting is manual. They take place in large facilities outdoors (e.g. chemical plants), so tight flight planning is unnecessary\footnote{{\tt www.cnet.com/roadshow/news/ford-dagenham
-plant-drone-safety/} and {\tt https://www.thedrive.com/
tech/22128/pilsner-urquell-used-flyabilitys-elios-
drone-to-inspect-beer-bottling-plant}, November 2020.}. Intelligent flight scheduling, nonetheless, may be needed to track indoor processes with a high degree of automation, and this problem has already attracted the attention of the research community \cite{Khosiawan2016,Khosiawan2018}. 

Quality control procedures are akin to surveillance applications. They are quite repetitive and, therefore, the resources they need can be scheduled. Particularly, in robotic plants drone flight scheduling is feasible even if drones must coexist with mobile robots, since they can avoid each other, as the movements of the robots are predictable. In these procedures drone communications may rely on the robust wireless networks \cite{Maghazei2019} that will be part of the Industry 4.0 paradigm \cite{Lasi2014}, and, even though the image recognition algorithms involved differ from those in surveillance, they can also be delegated to external servers in the plant. In general, it is expected that, with the advent of low latency 5G communications, edge computing \cite{Porambage2018} will enable many industrial use cases \cite{Chen2018}, so that computational offloading from the drones will not be an issue even if manufacturing plants themselves lack local computing resources. 

\begin{figure*}[!htbp]
\centering
\includegraphics[width=0.75\textwidth]{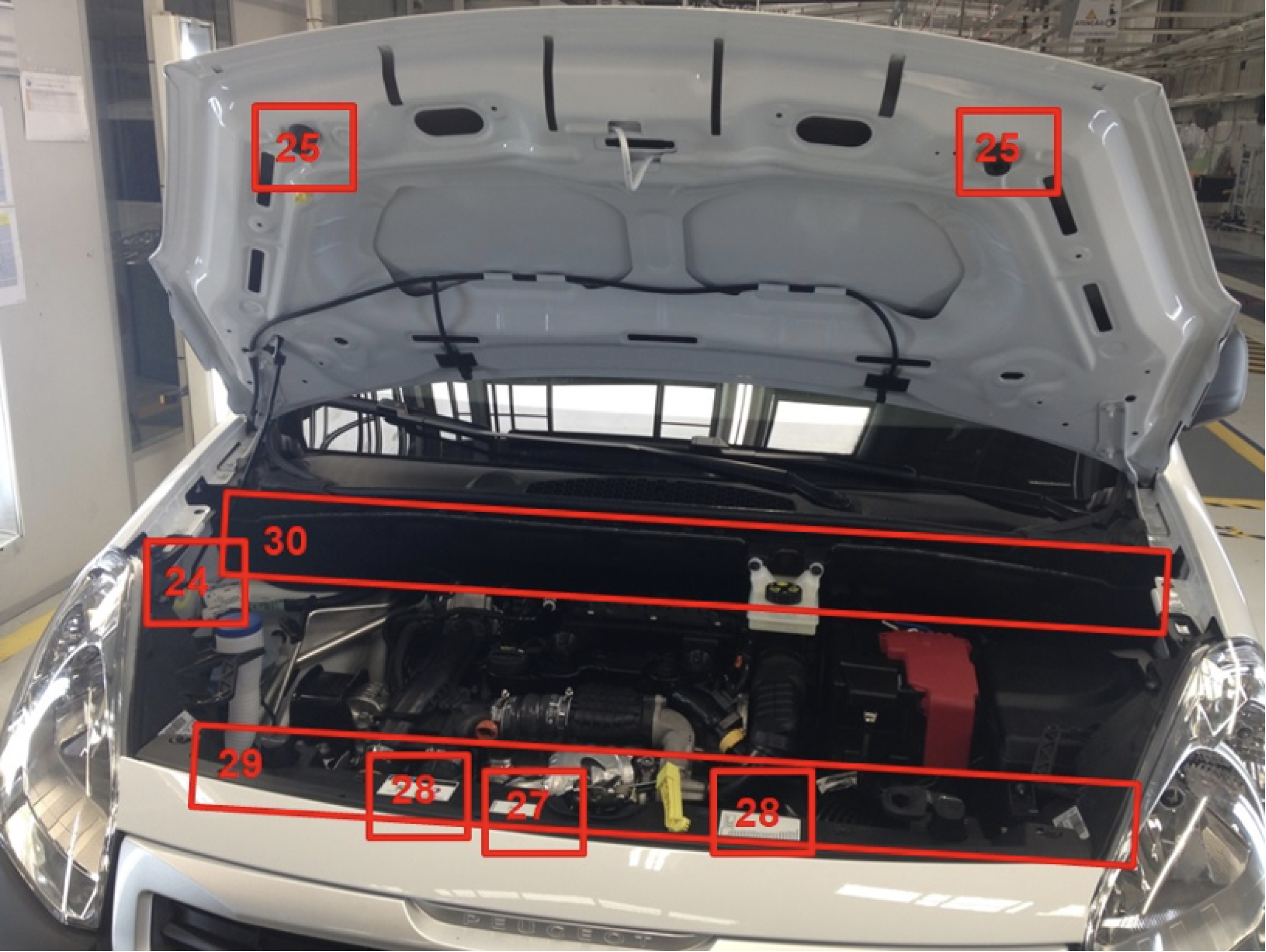}
\caption{\label{car}Quality control protocol of a car in a production chain (courtesy PSA Mangualde Portugal). The elements in the sequential protocol are highlighted in red.}
\end{figure*}

Diverse practical solutions for indoor airborne sensing have been studied. For example, in \cite{Wu2017} ultrasound sensors assisted piloting inside industrial facilities with limited line of sight. PSA, a major car manufacturer, has considered airborne sensors for image recognition in production chains (see Figure \ref{car}). Drones are much more flexible for this type of environment than fixed cameras, because drones can easily adapt themselves to production line rescheduling and they can coexist with predictable dynamic obstacles in robotic areas. In other words, a key differential characteristic of the indoor quality control scenario (unlike outdoor surveillance) is that drone flights are deterministic, so they can be automated. 

Figure \ref{car} shows a car that has been prepared for a manual quality check in a real production line. The red boxes highlight the areas in which a human operator must check information stickers and plastic parts, by following a written protocol (note the protocol sequence numbers in the red boxes). The idea is replacing the human operator by a drone that will circle the car while transmitting images of those areas to an image recognition server. In the case in the figure, considering that the drone must place itself at spots at the right distance for its camera field, six stopping points should be necessary (front, left and two at each side of the vehicle). The narrowest free space around the vehicle in the manipulation cage is around one meter wide, leaving around forty centimetres at each side of the drone when it crosses that space. The flight plan of the drone depends on the particular model of the car that occupies the manipulation cage, but this information, as well as the time the vehicle will spend in the cage, are known beforehand. A group of drones departing from charging stations can cover the whole production line.

In an indoor scenario GPS will typically be useless. Low cost distance measurement sensors such as ultrasound sensors or Light Detection and Ranging (LIDAR) Time of Flight (ToF) rangefinders can be used to avoid collisions, but the main positioning technology should have a longer indoor range (around 10 m) and provide $\sim5$ cm accuracy during flight (it is assumed that the charging station will have its own close range positioning solution, by relying on image recognition or other precision landing methods such as infrared beacons).

\section{BACKGROUND}
\label{RelatedWork}

 There exist diverse positioning alternatives that fit into the requirements in the previous section. We decided to focus on two candidates, visual odometers and Ultra-Wide Band (UWB) positioning, because they need few or no references, the equipment that must be installed in the drone is light, and they can be easily deployed in a technical environment (in theory visual odometers need no references as they simply process variations in background images, whereas UWB positioning needs three or four radio beacons with no wireless backbone for triangulation purposes). Also, there exist robust commercial implementations. For example, Intel released in 2019 the RealSense T265 Tracking Camera, a light and cheap device (less than \$200) with two fisheye lens sensors, an Inertial Measurement Unit (IMU) and a Visual Processing Unit (VPU), which executes a visual Simultaneous Localization and Mapping (SLAM) algorithm \cite{Mise2020}. Thus, this device is an adequate visual odometer for small drones. Regarding UWB indoor localization systems, Decawave's transceivers are specifically designed for accurate indoor positioning \cite{Ruiz2017}. Pozyx integrates Decawave transceivers in beacons and tags that determine distances and orientations respect to UWB beacons \cite{Kulmer2017}. 

Vision based methods are nowadays widely employed in unmanned autonomous vehicle navigation \cite{Kanellakis2017,Lu2018,Sanchez-Rodriguez2018}. UWB localization has been applied to this field as well \cite{Macoir2019}. Indeed, UWB has good propagation characteristics in indoor industrial facilities. In \cite{Schmidt2018Experimental} it was shown that it is not only useful for localization purposes, but also for industrial communications, whereas technologies like WiFi and ZigBee do not meet certain requirements of data rate, power consumption and robustness. 

The well known Kalman algorithm is a common sensor fusion approach that estimates the state of a system by combining sequential measurements provided by different sensor technologies \cite{Yukun2007}. A Kalman filter is a recursive least mean square algorithm that calculates the next state of a dynamic system by assuming a Gaussian distribution of noisy observations. It has been widely used for optimal control of navigation systems since its conception, and different models have been applied to unmanned aerial vehicle localization \cite{Mao2007} and tracking \cite{Schubert2008}. The Constant Turn Rate and Acceleration (CTRA) model \cite{Polychronopoulos2004} considers the motion clothoid of the target. It yields better performance than other state-of-the-art models such as the Constant Velocity (CV) and Constant Acceleration (CA) models for general motion tracking \cite{Khalil2016}. It assumes constant turn rate and tangential acceleration of the target. Notwithstanding these assumptions, this approximation of real motion is adequate for the scenario in our research (drone motion tracking around a car with minimal height variations).

In \cite{Zeng2019} the authors proposed a Kalman fusion method that exploits the different intrinsic advantages of visual odometers and UWB positioning. In particular, they stated that visual odometry can smooth out UWB measurement data (which are much noisier) and compensate for the deficiencies caused by multipath propagation. They also stated that UWB sensors can correct the cumulative error produced by visual odometry. Their approach simply applies Kalman filtering to linear combinations of the outputs from the two sensor types, and it yields satisfactory performance. Therefore they gain in simplicity by ignoring the statistical dependencies between sensor readings, unlike more complex fusion schemas (see \cite{Javadi2020} for a comprehensive review). In general, sensor fusion is advantageous compared to the independent usage of any of the technologies alone (consider for example the fusion of UWB and micro electromechanical IMU information \cite{Corrales2008, Sirtkaya2013}), if certain conditions are met. In our production chain scenario, flight plans will be expected to be short and noncomplex. In these plans, drones will perform small hops between predetermined stopping points, and they will return to their bases or charging stations periodically. The main problems will be: (i) achieving enough position accuracy at certain stopping points (from which quality control images will be taken and transmitted to the application server) and (ii) guaranteeing a straight flight in the segments between those stopping points (while maximizing the distance to known obstacles). We found that visual odometers excel in the second goal but they may return incorrect outputs if their visual references get lost for any reason (so that cumulative errors get exacerbated). Therefore, we propose a self-corrective approach in which filtered and clustered UWB readings provide references to estimate the stopping points, for correcting the underestimations of the (much less noisy) readings of visual odometers. We demonstrate that this approach outperforms the Kalman fusion in \cite{Zeng2019} when the mutual error between the technologies involved becomes too large.

Let us remark that other authors have already studied the problem of incorrect airborne sensor readings. For example, R. Wang et al. analyzed with a Hidden Markov Model the “digital upsets” in airplane sensors due to electromagnetic interference \cite{Wang2017}. We also indirectly detect sensor disturbances (in our case in visual odometry), although we do so by comparing the estimates from sensors of different technologies. Z. Zhao et al. have proved that by analyzing flight dynamics it is possible to detect sensor anomalies and performance degradation \cite{ZHAO2019}. In particular, they considered different “modes” corresponding to different faulty sensors, and they assumed that two different sensors cannot fail simultaneously (in our scenario this is also empirically observed). A main difference with our scenario is that our method corrects the estimates from a faulty sensor (the visual odometer) using the estimates from another sensor (the UWB system), by generating cumulative correction vectors.

\section{Hardware Design}
\label{materials}
\begin{figure*}[!htbp]
\centering
\includegraphics[width=0.80\textwidth]{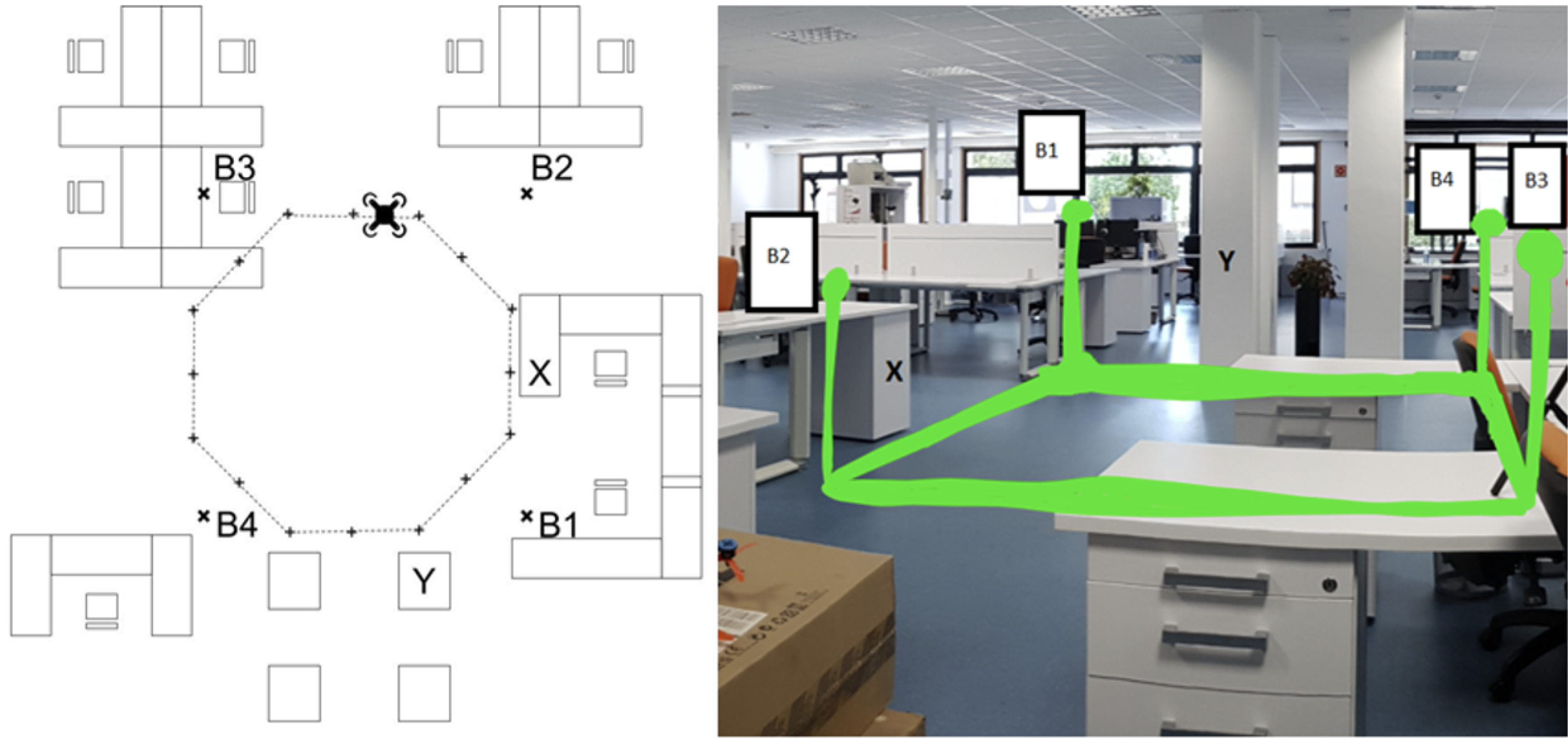}
\caption{\label{drone_movement}Laboratory testbed indicating the locations of Pozyx anchors B1 - B4, surface X and reinforced concrete columns Y.}
\end{figure*}

We deployed in our lab a Pozyx positioning system with four anchors B1 - B4 in a square arrangement as shown in Figure \ref{drone_movement}, all of them at a height of 1 m. The Pozyx unit provided absolute localization data based on trilateration in a coordinate system, as described in \cite{Zeng2019}. It is possible to improve its accuracy by combining multilateration with anchor selection to overcome measurement errors in extreme environments \cite{Mimoune2019}, but this was not our case, so we employed the default algorithm settings. The system has a bandwidth of 500 MHz with 0.16 ns pulses. It has been used for industrial indoor positioning \cite{Barral2019}, urban navigation \cite{Moran2018} and photogrammetry \cite{Masiero2017} research, just to cite some examples.

\begin{figure}[!htbp]
\centering
\includegraphics[width=0.30\textwidth]{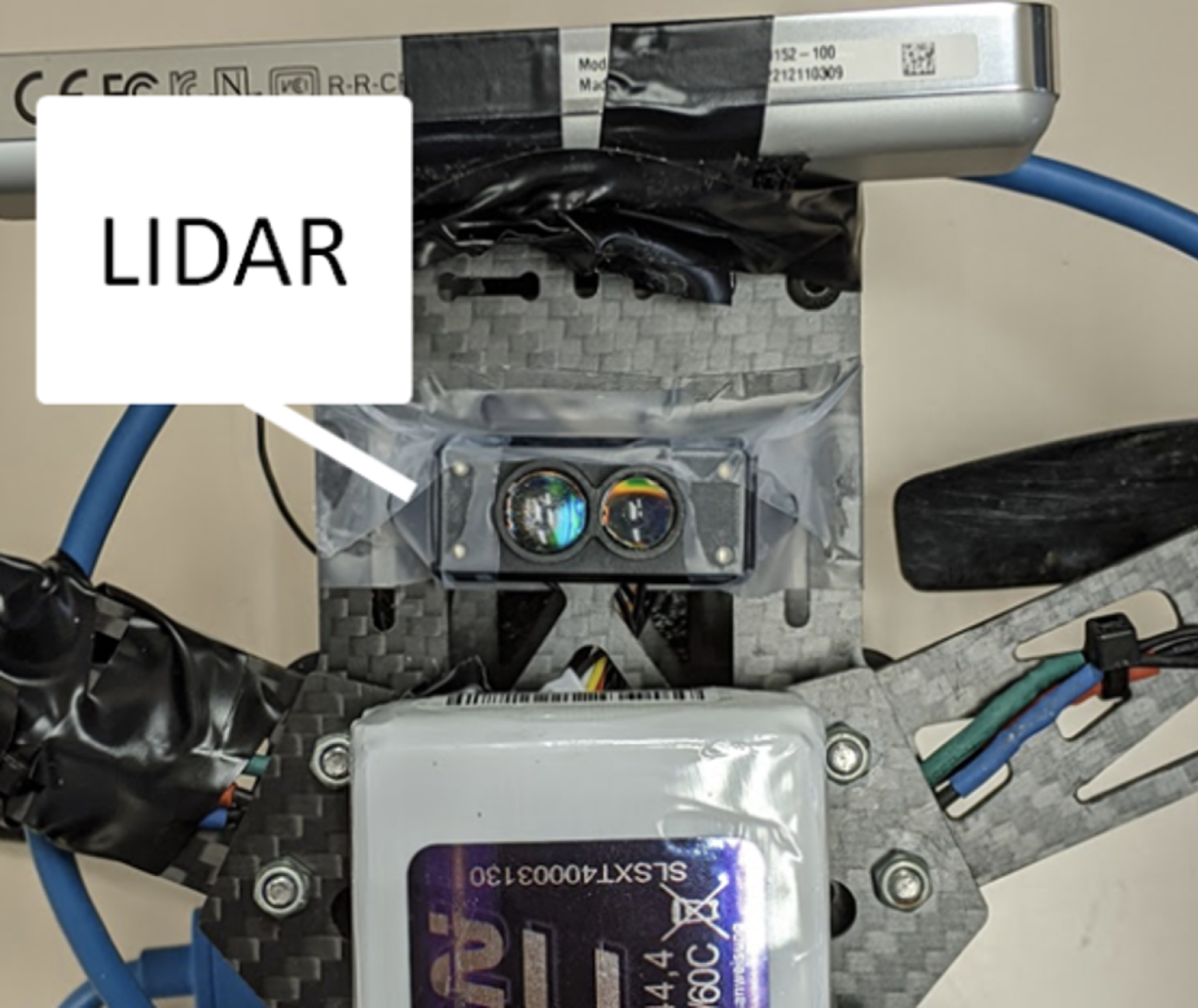}
\caption{\label{drone}Drone LIDAR.}
\end{figure}

Pozyx anchors are static. Their installation can be planned or, alternatively, the anchors can calibrate themselves, for which the manufacturer provides a software application. We chose the first option because it was more precise. The square layout of the anchors was measured with a Bosch GLM30 laser telemeter and the resulting parameters were uploaded to the Pozyx monitoring software in the control station. Measurement data are extracted from Pozyx tags via a serial connection. The user can select the exact data to be transferred. For our experiments we chose $x$, $y$ and $z$ coordinates with timestamps, although the $z$ coordinate was discarded as the drone could take exact height measurements by pointing a LIDAR to the ground (Figure \ref{drone}).

\begin{figure*}[!htbp]
\centering
\includegraphics[width=0.80\textwidth]{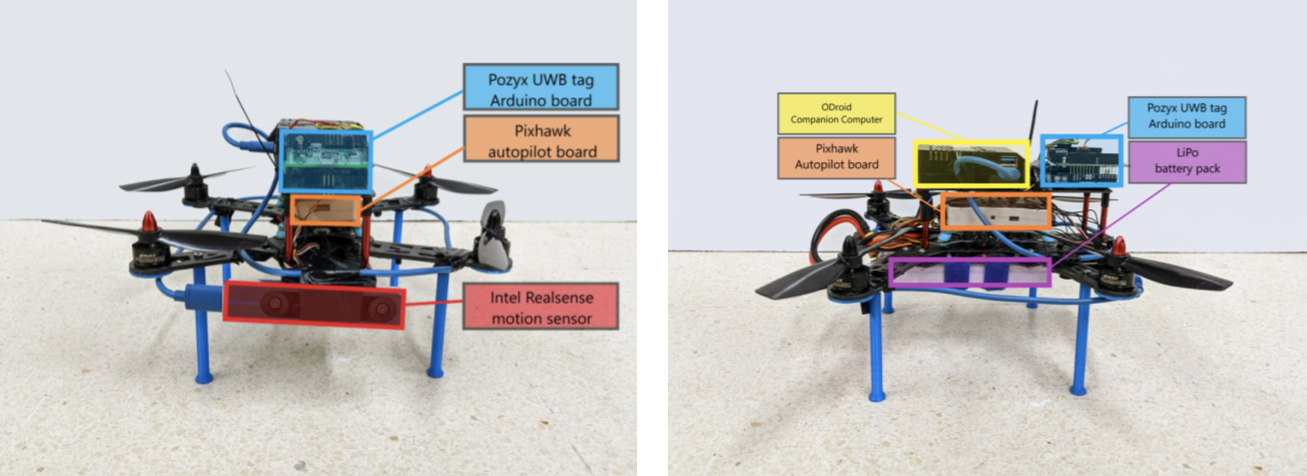}
\caption{\label{drone_sensor}Drone sensor platform. Left: front view. Right: rear view.}
\end{figure*}

The visual odometer was an Intel RealSense T265 unit. As previously said, this is not a pure visual odometer. Besides of two fisheye lenses, it also includes IMUs and an integrated VPU that runs a SLAM algorithm. This light device (55 g) with low battery consumption (300 mAh) can be easily attached to a small size drone. In \cite{Carfagni2019}, the authors characterized it. They reported that it is sufficiently accurate to acquire information at very close range (from 15 cm to 50 cm), so it is adequate for our purposes. Unlike the Pozyx unit, the RealSense device does not need any configuration. Its firmware allows it working out of the box. For accessing RealSense T265 data, there exist a driver and an API with interfaces for many programming languages. Our choice was Python for its simplicity. Again, the data we extracted were $x$, $y$ and $z$ coordinates with timestamps, and we also discarded $z$. 

Unlike the Pozyx unit, the RealSense T265 unit does not depend on external peripherals. Its tracking calculations are based primarily on information gathered from the two onboard fish eye cameras, each with a 160 degree field of view, capturing 30 frames per second. This wide field allows keeping points of reference visible for relatively long times as the drone passes by. The images from the visual sensors are combined with data from the onboard IMU and they jointly feed the proprietary localization algorithms. The system has been used for medical \cite{Siena2018} and robotic mapping \cite{Ahn2019} research, just to cite some examples.

The controller extracted data simultaneously from the interfaces of both sensors with a Python script. As the RealSense T265 device works at a higher measurement rate, the timestamps were matched to preserve the temporal consistency of the data. Also, a trivial conversion was applied to the coordinates, because they were expressed in different units (mm and cm).

Regarding their references, the Pozyx system requires three anchors arranged in two perpendicular straight lines as the $x$ and $y$ axes of its coordinate system. A fourth anchor can be placed anywhere and is not required to form a perfect rectangle with the other three. Coordinate (0, 0, 0) is the precise location of the first anchor, which has always the lowest identifier. The RealSense T265 device sets coordinate (0, 0, 0) as the precise location where the system starts recording data after booting, and the $x$, $y$, and $z$ axes are set depending on the initial motion of the camera. 

Figure \ref{drone_sensor} shows the drone we developed, a custom made carbon fiber platform with four 2300 KV brushless motors, 30 A electronic speed controllers, a 3S 4000 mAh battery and a Pixhawk 2.4.8 autopilot board. This configuration weighs less than 700 g without any payload. Once the 359 g battery is installed, the empty weight of the setup is around one kilogram. Its thrust is enough for lifting a payload consisting of an Odroid XU4Q companion computer, the RealSense T265 device and the Pozyx tag, which weigh 38, 55 and 12 g respectively. Overall flight time is around 12 minutes, at a worst-case battery discharge of approximately 80\%. In case of needing extra thrust, the propeller layout can be modified. The propellers in Figure \ref{drone_sensor} were chosen for maximizing drone stability, with indoor image acquisition in mind.

This platform can support a wide spectrum of applications satisfying the payload limits. Its excellent stability allows for safe and controlled indoor flights avoiding the obstacles in the manipulation cage. Regarding data transmission, there also exist diverse alternatives. In addition to a 433 MHz transceiver for serial telemetry and automated flight commands, the onboard computer supports Wi-Fi and LTE dongles that can be installed for extended wireless communication. 

\section{PROPOSED METHOD}
\label{methods}
The terminology in this section is as follows: we wish to estimate a sequence of position vectors $y[n] \in R^{2}$ from some noisy estimations $x_u[n] \in R^{2}$ and $x_o[n] \in R^{2}$, corresponding respectively to UWB triangulation and visual odometry, both related to the same time index $n \in {\{0,\cdots,N\}}$.

\subsection{EXPERIMENTAL MOTIVATION}
\begin{figure*}[!htbp]
\centering
\includegraphics[width=0.80\textwidth]{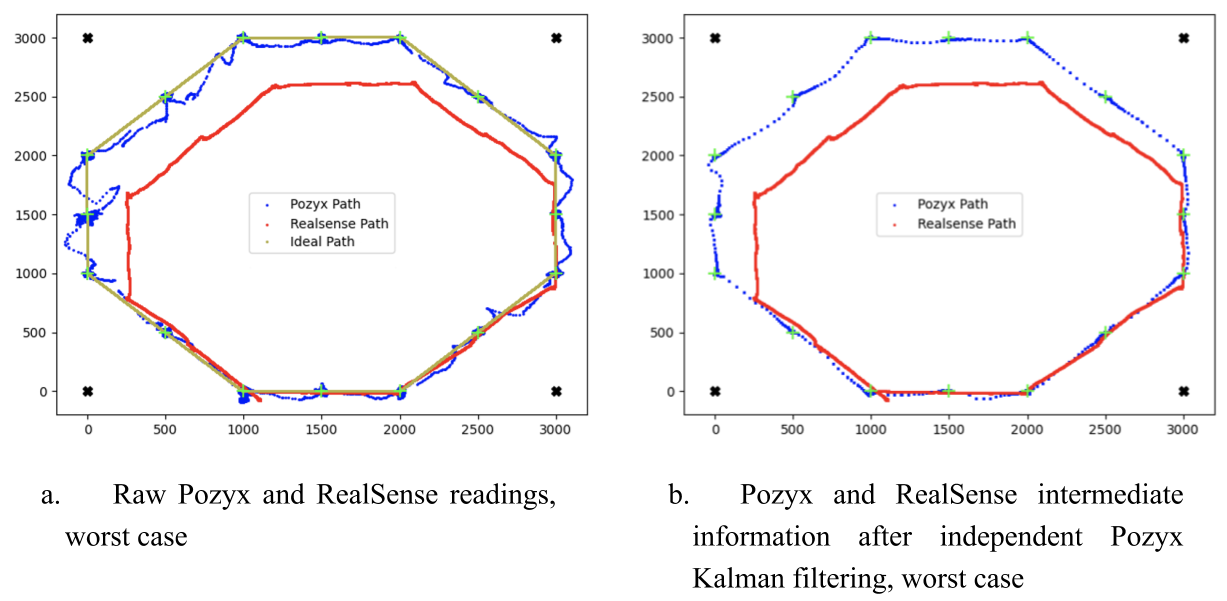}
\caption{\label{worst_case}Worst case in the laboratory testbed with severe loss of RealSense references. Stopping points marked as “+”. Pozyx anchors marked as “x”. Ideal path marked as straight segments. Scales in mm.}
\end{figure*}

Figure \ref{worst_case}.a shows an example of raw readings of the RealSense T265 and Pozyx sensors for respective sampling rates of 200 Hz and 27 Hz, along an octagonal trajectory inside the green square in Figure \ref{drone_movement}, where the first stopping point was (1000, 0) and the drone moved counterclockwise until completing a cycle. This example corresponds to the {\it worst case} in our experiments with severe loss of RealSense references around the sixth stopping point (3000, 1500). The positions of the Pozyx beacons on the floor plane are marked with “×” symbols. Beacons B1, B2, B3, and B4 were thus placed at coordinates (3000, 0), (3000, 3000), (0, 3000) and (0, 0), respectively. As previously said, the sensors were installed on the drone platform in Figure \ref{drone_sensor}, which flied straightly in automated mode between each successive pair of stopping points (therefore, the measurements between stopping points should ideally reflect straight lines). There were 16 such points, marked with “+” symbols in Figure \ref{worst_case}. 

As shown in Figure \ref{worst_case}.a, raw UWB readings $x_u[n]$ were in general much noisier and more irregular than those of the RealSense unit, $x_o[n]$, even after applying built-in Pozyx default enhancements. These readings can be affected by obstacles and metallic elements. Note for example the irregular pattern between stopping points (0, 2000) and (0, 1000), which seemed due to the metallic equipment in the workstations to the right of Figure \ref{drone_movement} (not shown in the image). Note also the columns marked with an “Y” symbol in Figure \ref{drone_movement}, which are made of reinforced concrete and thus could reflect signal energy. Regarding the visual odometer, it severely underestimated the length of some displacements between stopping points in 40\% of the trials, possibly due to varying illumination conditions. Consider the case between stopping points (3000, 1000) and (3000, 2000) in the example in Figure \ref{worst_case}.a, for instance. This behavior was clearly associated to the uniform surface with no contrast at all marked with an “X” symbol in Figure \ref{drone_movement} (since the RealSense T265 cameras pointed to that direction when the drone was moving nearby). Regarding the respective advantages of the two technologies during their initial assessment, UWB positioning, despite of its noisier output, was able to register measurements centered on all stopping points, whereas the visual odometer registered rather linear trajectories in between, regardless of their (sometimes wrongly) estimated length.

Let us remark that it is possible to mitigate the issues of both methods to some extent: by separating UWB beacons from troublesome spots, so that their transmissions experience less blockage and reflections; and by placing stickers with rich visual references on uniform surfaces. Nevertheless, in industrial environments it could be difficult to find fully obstacle-free positions and visual references might get scratched and fade over time. Therefore, it will still be interesting to combine different technologies to increase the probability that at least one of them works properly.

\subsection{SELF-CORRECTIVE APPROACH}
In this work we propose the following self-corrective approach, with three method components: A) independent Kalman filtering of UWB data (to avoid the effect of high mutual errors in Kalman fusion), B) data association by means of stream clustering (to filter out UWB noise at stopping points) and C) Correction of odometer data with filtered UWB data based on the generation of cumulative vectors (when sensor readings diverge).

\begin{enumerate}[label=\Alph*)]
\item {\it Independent Kalman filtering}. Here, by mutual instantaneous positioning error we refer to $d(x_u[n], x_o[n])$, where $d(\cdot)= \| x_u[n] - x_o[n] \|_2$ is the Euclidean distance. If this error becomes too large, it will be likely that one of the technologies has failed temporarily, so its outputs could be problematic for Kalman fusion, for example. In particular, in our scenario, visual odometer errors may become too large if the unit underestimates the displacement. Thus, unlike in \cite{Zeng2019}, we apply Kalman filtering independently to Pozyx readings $x_u[n]$ to obtain intermediate information $y_u[n]$ (RealSense T265 readings are so clean that we took $y_o[n]= x_o[n]$ directly). Regarding the Kalman filter model, we obtained satisfactory results with the CTRA model we mentioned in Section \ref{RelatedWork}. Specifically, we followed a freely available implementation\footnote{Available at {\tt https://github.com/balzer82/Kalman/
blob/master/Extended-Kalman-Filter-CTRA.ipynb}, November 2020.}, by adjusting the covariance matrices to our scenario as described in Section \ref{results}. Figure \ref{worst_case}.b shows the RealSense T265 and Pozyx intermediate signals $y_o[n]$ and $y_u[n]$ corresponding to the example in Figure \ref{worst_case}.a once independently processed (note how Kalman filtering smooths out the zig-zag pattern between (0, 2000) and (0, 1000) in Figure \ref{worst_case}.a). 

We next describe the particularization of the Kalman CTRA model, an Extended Kalman Filter (EKF), to our case. As its name indicates, it assumes that the turn rate and the acceleration of the drone are constant. Our scenario fits well into these assumptions, because the drone adjusts its direction between trajectory segments by rotating around the $z$ axis and accelerates between stopping points (for this reason, a constant velocity model is not valid).

The state of the system is defined by $x_{k}= (x,y,\upsilon,\psi,\psi^{\prime},a)$, where the $x$ and $y$ coordinates indicate the drone position, $\upsilon$ is the linear velocity, $\psi$ is the heading direction angle, $\psi^{\prime}$ is the yaw rate and $a$ is the acceleration, where the last two are supposed to be constant. The predicted state is:
\begin{multline}
x_{k+1} = g(x,y,\upsilon,\psi,\psi^{\prime},a) = \\
(x + \frac{v}{\psi^{\prime}}(-sin (\psi) + 
sin(T\psi^{\prime}+\psi)), \\ y +\frac{v}{\psi^{\prime}}(cos(\psi) - cos(T\psi^{\prime}+\psi)), \\ T\psi^{\prime}+\psi, Ta+v, \psi^{\prime}, a)
\label{equation1}
\end{multline}

Where $T$ is the sampling period. The rest of the EKF equations are:

\begin{itemize}
	\item Projection of error covariance: $P_{k+1}=J_{k} P_{k} J_{k}^{T} + Q$
	\item Kalman gain: $K_k = P_k (P_k + R)^{-1}$
	\item State update via measurements $u_k$: $x_k = x_k + K_k (u_k - x_k)$
	\item Error covariance update: $P_k = (I - K_k) P_k$
\end{itemize}

All matrices have dimension 6×6 in our case. $J_k$ is the Jacobian matrix of $g()$ with respect to the state vector, $Q$ is the process noise covariance diagonal matrix and $R$ is the measurement noise covariance matrix. All measurements $u_k$ are calculated from positioning data, although $a$ can be taken from the drone accelerometers.

The Kalman filter is restarted at each stopping point. In Section \ref{results} we detail all parameter settings and variable initializations.

\item {\it Denoising of UWB positioning data at stopping points}. UWB data $y_u[n]$ form noisy clouds around these points (see Figure \ref{worst_case}.a). As noted in \cite{Javadi2020}, even though the Kalman filter and its derivatives are adequate for estimating the positions of the targets, complementary data association algorithms are useful for identifying targets (in our case the next stopping point). This is the motivation for method component B. To eliminate the noise we apply an unsupervised clustering algorithm to estimate the stopping points as the centers of clusters of measurements. Specifically, inspired by the Density Based Clustering method \cite{Kriegel2011} we formulated the stream clustering Algorithm \ref{algorithm1}, which is activated within Euclidean distance $\gamma$ from each stopping point $s_i$:

\begingroup
\newlength{\xfigwd}
\setlength{\xfigwd}{\textwidth}
\begin{lstlisting}[caption={\vspace{\hcaption}Component B of the method.}, label={algorithm1},mathescape=true,basicstyle=\small]
Initialization: $L = \emptyset, C = \emptyset, candidate = 0$
Repeat:
 If $y_u[n] \notin L$ then $c(y_u[n])$ = 0 and $y_u[n] \rightarrow L$.
 $ \forall z \in L$, if $z \neq y_u[n]$ and $d(z, y_u[n]) \leq \alpha$ then 
 $c(y_u[n]) = c(y_u[n]) + 1$ and $c(z) = c(z) + 1$.

 $ \forall z \in L$,
 If $c(z) \geq K_1$ then 
 $z \rightarrow C$ and $candidate$ = 1
 If $c(z) = K_2$ and $candidate$ = 1 then
 end repeat
 
$s_i^{\prime} = argmax(c(z))$, $z \in C$
\end{lstlisting}
\endgroup

where $L$ is a temporary list of Kalman output vectors, $z$ is an element of $L$, $c(z)$ is an auxiliary counter (one per each element $z$ in $L$),
parameter $\alpha$ is the maximum intracluster distance, $C$ is the temporary set of denoised candidates for predicting stopping point $s_i$, parameter $K_1$ is the number of neighbors above which a cluster is suspected to exist around a point, and parameter $K_2$ is the number of neighbors at which the current algorithm instance is terminated and an estimate $s_i^{\prime}$ for the $i-th$ stopping point is provided. Logically, the values of $\alpha$ and $K_1$ must be tuned to maximize the elimination of outliers around a stopping point. If these parameters are too large, the method may lose denoising performance. The method works because two samples in close vicinity will have similar neighbor populations. The two different parameters $K_1$ and $K_2$ thus define a hysteresis region.

In some repetitions of our experiments, the drone experienced intermittent connections with some UWB anchors at some stopping points. Consider the example of the stopping point in Figure \ref{symmetric_out}. Note the main “cloud” of Pozyx readings around the stopping point and the linear pattern of outliers that extend nonlinearly to the right when one of the anchors becomes temporarily unreliable, so that non-Gaussian non-convex readings emerge around the stopping point.

We remark that the density based association in method component B could be replaced by any other data association algorithm, such as the Nearest-Neighbor Standard Filter (NNSF) \cite{mcmillan1990data} or the Probabilistic Data Association Filter (PDAF) \cite{Bar-Shalom2009}. However, despite its simplicity, noisy readings around some stopping points, as in the example in Figure \ref{symmetric_out}, discouraged choosing NNSF, whereas PDAF assumes a Gaussian distribution of noise and convex noise clouds, which does not always hold in our scenario. For a comprehensive review of data association algorithms see \cite{Javadi2020}. Logically, in our case, the multi-target variants in \cite{Javadi2020} do not apply, since there is a single target (the next stopping point). 

\begin{figure*}[!htbp]
\centering
\includegraphics[width=0.55\textwidth]{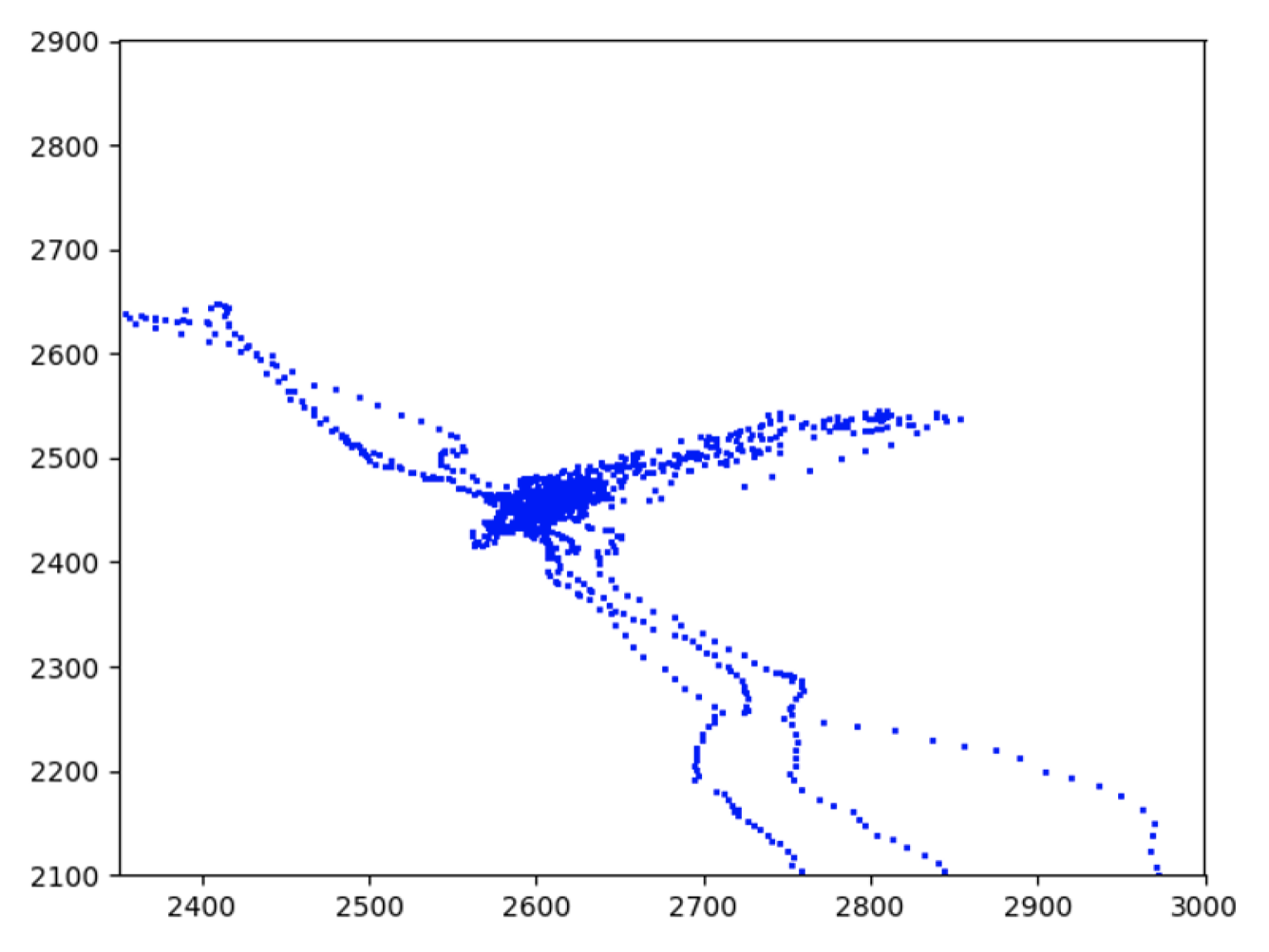}
\caption{\label{symmetric_out}Example of asymmetric outliers in the event of temporary loss of Pozyx anchor connection. Scales in mm.}
\end{figure*}

\item {\it Correction stage}. This is the method component that introduces the self-corrective combination of sensor outputs. It is based on the observation that visual odometer displacement errors are cumulative, as mentioned in \cite{Zeng2019}. Therefore, if we can correct a visual odometer error at some point, the same correction vector should be applied to all visual odometer measurements from that moment on. Basically, we trust Pozyx estimates at the stopping points. Let $S^{\prime}$ be the set of stopping point estimates that result from applying Algorithm 1 in method component B above to Pozyx outputs. Then, let $w_i$ be a correction vector (which we initialize to (0, 0)) that must be added to all visual odometer outputs $x_o[n]$ between stopping point estimates $s_i^{\prime}$ and $s_{i+1}^{\prime}$, $y_o[n]= x_o[n] + w_i$, and $s^{\prime}_i=argmin(d(y_o[n], s_i))$. This is applied unless the mutual error between technologies becomes too large, $d(y_o[n],y_u[n]) \geqslant \beta $, where $\beta$ is a mutual error threshold, in which case the visual odometer is suspected to have lost its references and the stopping point $s^{\prime}_i$ is detected by method component $B$. If, during the current trajectory, at some pair of associated points $s^{\prime}_i$ and $y_{o,i}$, $d(s^{\prime}_i, y_{o,i}) \geqslant \beta $, where $y_{o,i}$ is the closest vertex in $y_o[n]$ to $s{^\prime}_i$ ($y_{o,i}=argmin(d(y_o[n], s^{\prime}_i))$), then $w_i = w_i + s^{\prime}_i - y_{o,i}$, and the RealSense T265 unit is rebooted. Logically, we assume that the UWB positioning unit is able to detect all stopping points.
\end{enumerate}
 
\begin{figure*}[!htbp]
\centering
\includegraphics[width=0.80\textwidth]{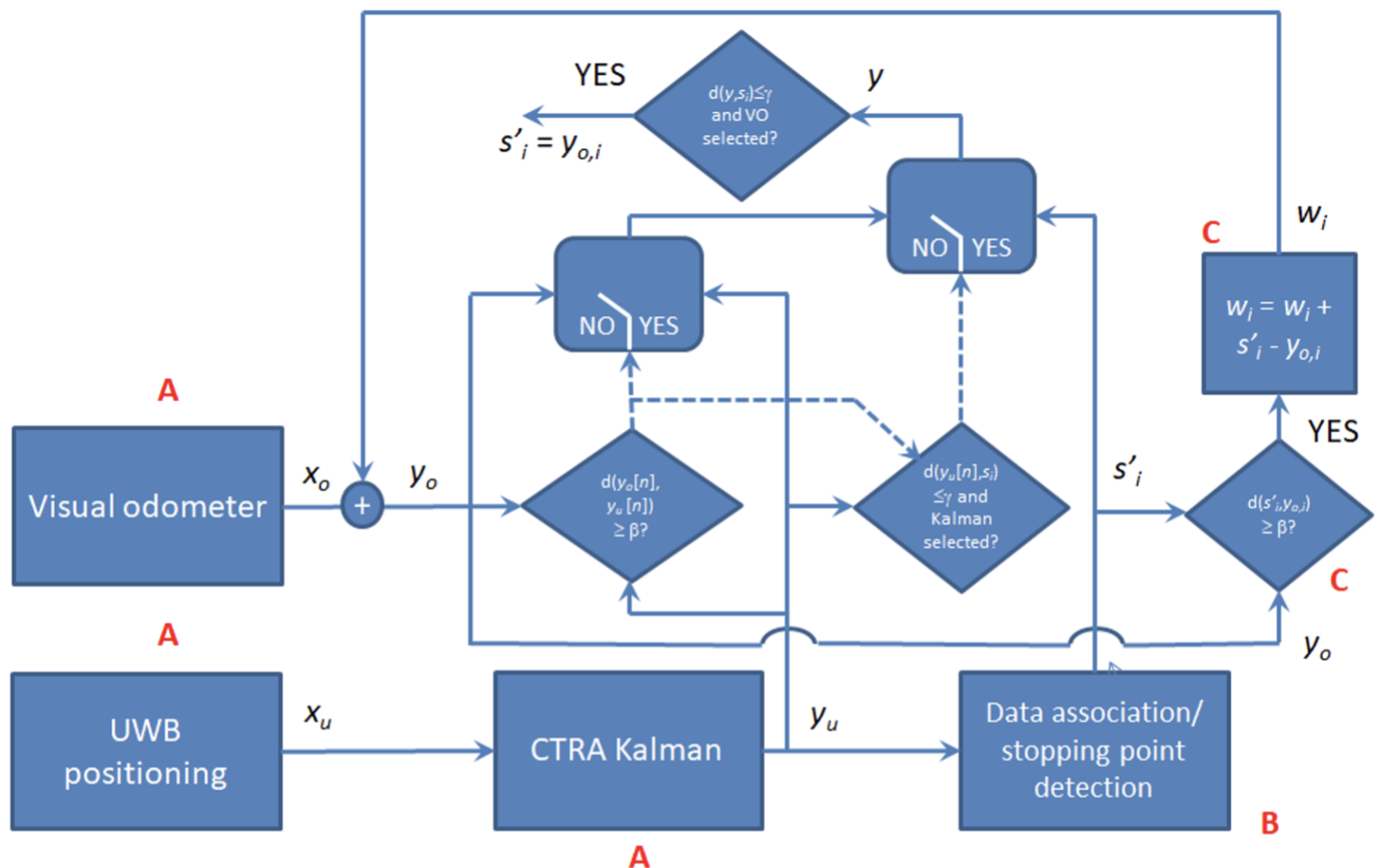}
\caption{\label{diagram}Flow diagram of the method. Red letters A-C indicate the method components (VO: visual odometer, “Kalman selected” condition holds when $d(y_o [n],y_u [n])\geq\beta$, “VO selected” condition holds otherwise).}
\end{figure*}

Figure \ref{diagram} formalizes the complete method as a flow diagram. We indicate with red letters the method components to which the different blocks belong. Note that method component B for stopping point detection is only needed in case the readings from the visual odometer diverge from UWB sensor readings. Otherwise, the visual odometer takes the drone to the stopping points with high accuracy, as shown in Section \ref{results}.

As a closing remark note that we are using the Euclidean distance instead of a probabilistic measurement like Mahalanobis distance. This is because all conditions $d(y_o[n],y_u[n]) \geq \beta$, $d(s^{\prime}_i, y_{o,i}) \geq \beta$, $d(y_o[n],s_i)\leq \gamma$ and $d(y_u[n],s_i)\leq \gamma$ are based on deterministic data ($s_i$), ``clean'' RealSense data ($y_o[n]$, $y_{o,i}$) or data with reduced uncertainty by Kalman filtering or data association ($y_u[n]$, $s^{\prime}_i$). Therefore we did not consider probabilistic distance measures necessary and employed Euclidean distances instead for simplicity.

\section{RESULTS}
\label{results}
We tested the self-corrective approach in the scenario in Figure \ref{drone_movement}. The parameter settings of the laboratory testbed were:
\begin{itemize}
	\item Pozyx sampling rate of 27 Hz 
	\item RealSense T265 sampling rate of 200 Hz
	\item Distance $\gamma$ was set to 100 mm
	\item Distance $\alpha$ was tuned between 5 and 20 mm in 5 mm steps
	\item Threshold $\beta$ was set to 3 cm and 6 cm in different tests.
\end{itemize}

Note that, even though the sampling rates of the two sensors are different, they are regular, so their readings can be synchronized for downsampling or interpolation. We refer the interested reader to \cite{Xue-Bo2012} as an approach to the problem of irregular sampling rates in target tracking.

In addition to our self-corrective approach, which includes CTRA Kalman filtering of Pozyx data, we considered pure RealSense tracking and other three Kalman variants in our tests:

\begin{itemize}
	\item CTRA Kalman filter applied only to Pozyx data. 
	\item Direct Kalman fusion: The same Kalman CTRA algorithm as in our approach, by assuming a single sensor that delivered both RealSense and Pozyx measurements, by merging RealSense and Pozyx measurements. 
	\item Kalman fusion: The same CTRA Kalman algorithm as in our approach, by assuming a single sensor that delivered $x[n] = \frac{1}{2} (x_o[n]+x_u[n])$ at Pozyx rate, so we synchronized RealSense measurements with Pozyx measurements as approximately as possible. This is similar to the approach in \cite{Zeng2019}, which obtained good results by applying Kalman filtering to linear combinations of the outputs of a visual odometer and a UWB positioning unit.
\end{itemize}

We tuned CTRA Kalman settings to maximize performance in our scenario. Specifically, we set $R=$ diag(25, 25, 0.01, 0.01, 0.01, 0.09) and $Q=$ diag(0.0625, 0.0625, 1.44e-4, 9e-6, 9e-4, 0.25) (in mm, radians and seconds where respectively applicable).

\begin{figure*}[!htbp]
\centering
\includegraphics[width=0.80\textwidth]{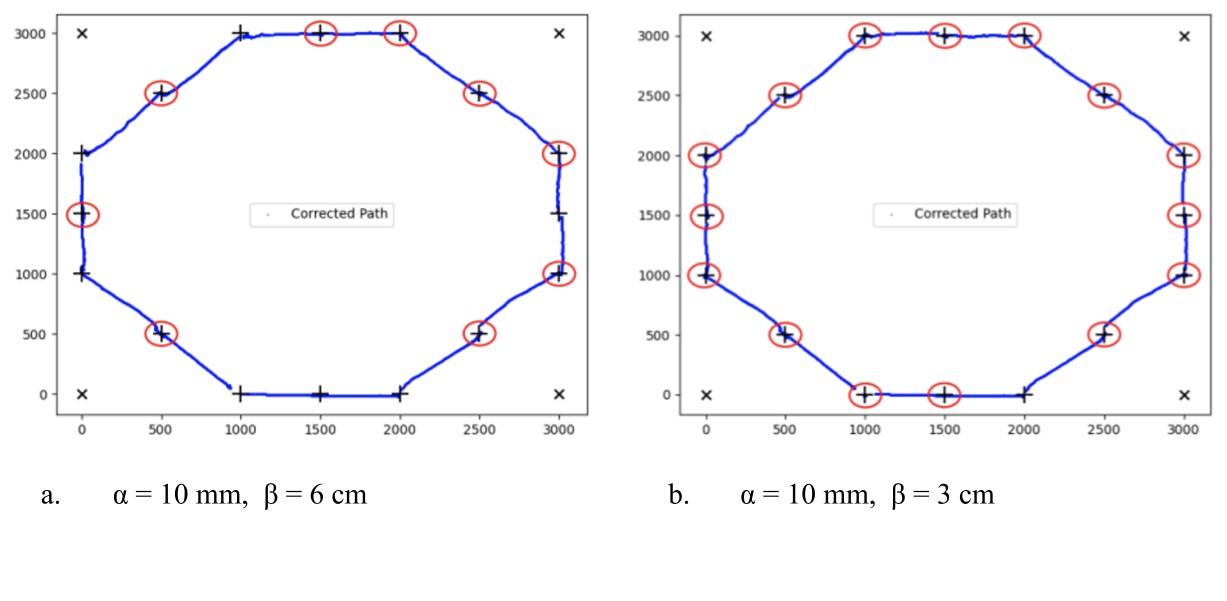}
\caption{\label{corrective}Self-corrective method, worst case. Scales in mm. Red circles indicate RealSense restarts.}
\end{figure*}

Parameters $\alpha$, $K_1$ and $K_2$ were set by successive trials for the two values of $\beta$ we considered, and we obtained the best positioning accuracy for $\alpha$ = 10 mm, $K_1$ = 100 and $K_2$ = 500 in our scenario. Figure \ref{corrective} shows the trajectories with our self-corrective approach for the worst case in Figure \ref{worst_case}, when the RealSense unit lost its references severely. Note that in that case it was necessary to update the correction vector in method component C (and thus to restart the RealSense T265 visual odometer) 9 out of 16 times for $\beta$ = 6 cm and 15 out of 16 times for $\beta$ = 3 cm. Therefore, for $\beta$ = 3 cm the roles of the two technologies were almost fully complementary, but for the stopping point at (2000, 0): in other words, in this case the approach automatically selected the RealSense T265 visual odometer for segment tracking and the Pozyx unit for stopping point detection and cumulative correction vector generation. 

\begin{figure*}[!htbp]
\centering
\includegraphics[width=0.80\textwidth]{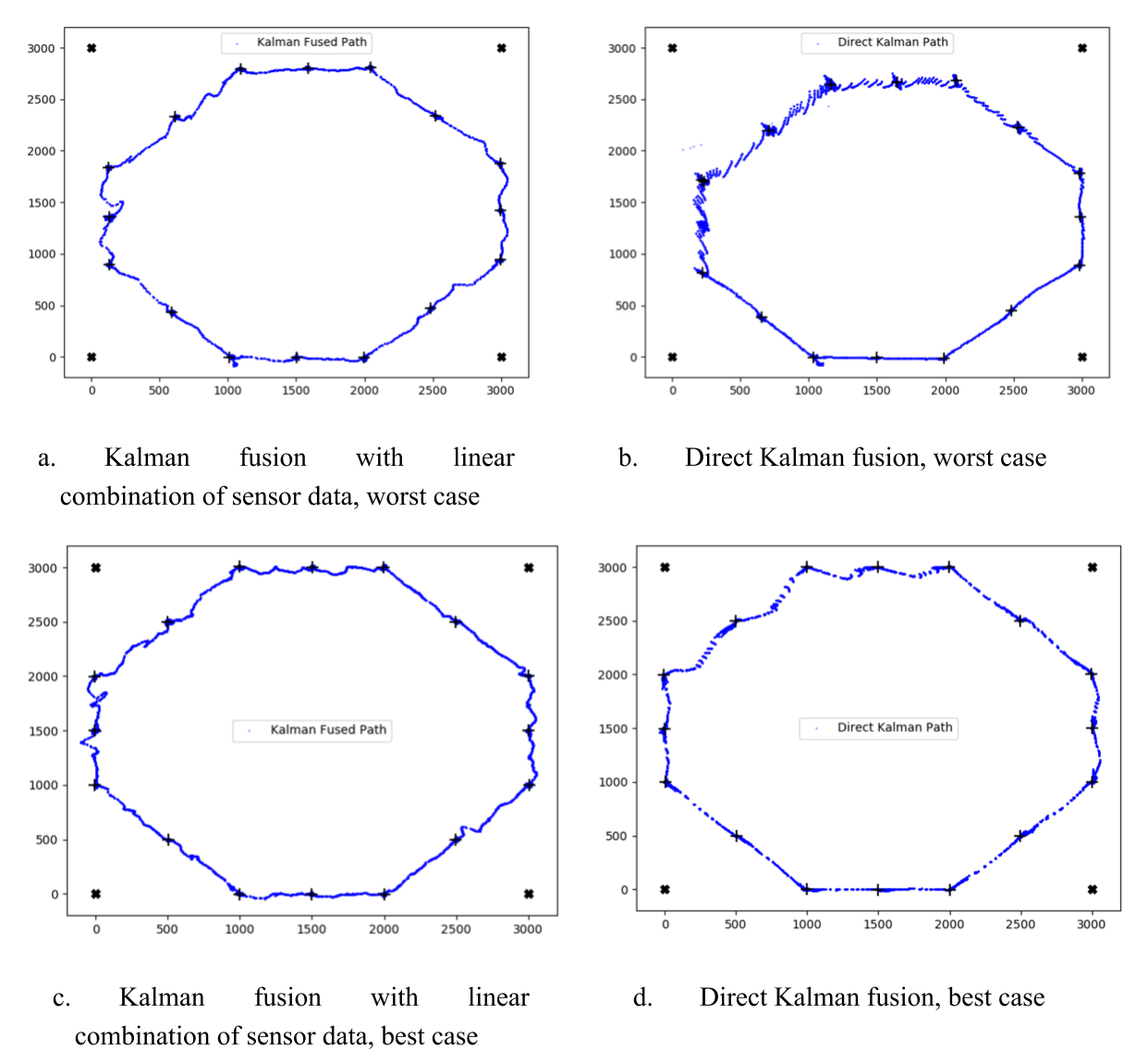}
\caption{\label{Kalman_fusions}Kalman fusions, worst and best cases. Scales in mm.}
\end{figure*}

Alternatively, we applied the two Kalman fusion variants we considered to the outputs of the two technologies. Figure \ref{Kalman_fusions} shows the results in the best (RealSense working correctly) and worst (RealSense losing its visual references severely) cases of the tests.

Observe that, in the worst case (RealSense data in Figure \ref{worst_case}), Direct CTRA Kalman fusion only worked reasonably well until the sixth stopping point (3000, 1500), up to which the visual odometer had satisfactory visual references. From that point on, the high deviation of the visual odometer had strong impact on detection accuracy and overall root mean square error (RMSE), as shown in Table \ref{tab:tolerances}. CTRA Kalman fusion based on linear combinations of sensor readings behaved better than Direct CTRA Kalman fusion in the worst case: until the sixth stopping point visual odometer data helped to reduce Pozyx noise and, from that point on, Pozyx data alleviated the large deviation of RealSense data. In the best case both Kalman fusion variants performed similarly.

Table \ref{tab:tolerances} summarizes average results for 10 separate tests. It shows the positioning accuracies (in mm) at the stopping points of our self-corrective approach and the three Kalman filter variants. Ground truth data was collected by placing the drone manually at the stopping points. The values for the self-corrective approach correspond to the worst case, as the differences among tests with this approach were negligible.

Therefore, our approach was comparable to the RealSense unit when the latter worked correctly, as expected given the algorithmic flow in Figure \ref{diagram}, which gives priority to RealSense intermediate data over Pozyx intermediate data if the mutual error is small. However, when the RealSense unit lost its visual references (as shown in Figure \ref{worst_case}), it was no longer able to provide sub-5-cm stopping point detection accuracy (observe the large standard deviation of the detection accuracy at the stopping points), unlike our approach. 

The results by only applying CTRA Kalman filtering to Pozyx data in laboratory conditions also met the accuracy goal by far, although they were worse in terms of stopping point detection accuracy than our self-corrective approach and the RealSense unit when the latter worked correctly. 

Direct CTRA Kalman fusion performed better than CTRA Kalman fusion based on linear combinations of sensor readings as in \cite{Zeng2019} if the RealSense unit worked correctly. Also in that case, the results of these two variants were superior to CTRA Kalman-filtered Pozyx data, reflecting the beneficial effect of RealSense data. However, if the RealSense unit failed both Kalman fusion variants could not satisfy the accuracy goal, and Direct CTRA Kalman fusion accused this issue more severely. In this situation our approach outperformed both of them. 

\begin{figure*}[!htbp]
\centering
\includegraphics[width=0.80\textwidth]{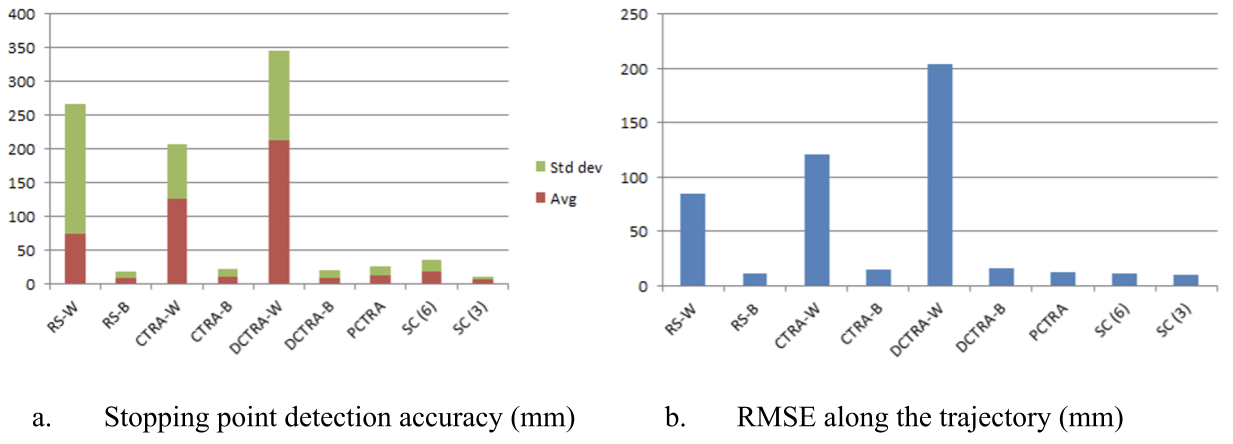}
\caption{\label{charts}Stopping point positioning accuracy (average and standard deviation) and trajectory RMSEs. RS: RealSense data, CTRA: CTRA Kalman fusion based on lineal combinations, DCTRA: Direct CTRA Kalman fusion, PCTRA: CTRA Kalman-filtered Pozyx data, SC(x): Self-Corrective approach with $\beta=x$ cm, W: Worst case, B: best case.}
\end{figure*}

Figure \ref{charts} offers a graphical view of the results in Table \ref{tab:tolerances}.

\begin{table*}[!htbp]
\centering
\caption{\label{tab:tolerances}Stopping point positioning accuracy (average and standard deviation) and RMSEs of the trajectories.}
\begin{tabular}{lllll}
\hline
 & & \begin{tabular}[c]{@{}l@{}}Avg. accuracy at \\ stopping points\end{tabular} & \begin{tabular}[c]{@{}l@{}}St. dev. accuracy at\\ stopping points\end{tabular} & \begin{tabular}[c]{@{}l@{}}RMSE along the \\ whole trajectory\end{tabular} \\\toprule
 
\multirow{2}{*}{\bf Self-corrective method} & $\beta$ = 3 cm, worst case & 5.13 mm & 4.52 mm & 9.51 mm \\
 & $\beta$ = 6 cm, worst case & 16.67 mm & 18.06 mm & 10.58 mm \\\hline
 
\multirow{2}{*}{CTRA Kalman fusion} & Best case & 9.25 mm & 12.26 mm & 14.24 mm \\
 & Worst case & 125.80 mm & 80.16 mm & 121.32 mm \\\hline
 
\multirow{2}{*}{Direct CTRA Kalman fusion} & Best case & 7.92 mm & 11,27 mm & 15.87 mm \\
 & Worst case & 212.04 mm & 134.06 mm & 203.58 mm \\\hline
 
\multirow{2}{*}{RealSense data} & Best case & 7.25 mm & 9,8 mm & 10.61 mm \\
 & Worst case & 72.64 mm & 193.07 mm & 84.55 mm \\\hline
 
Raw Pozyx data & Average & 131.90 mm & 47.59 mm & 76.50 mm \\ \hline

CTRA Kalman-filtered Pozyx data & Average & 10.87 mm & 14.9 mm & 12.7 mm \\ \bottomrule 
\end{tabular}
\end{table*}

\section{CONCLUSIONS}
The results of the experiments were consistent with our assumptions about the characteristics of indoor facilities. The challenges they pose to positioning technologies are less evident in more open scenarios like the experimental setting in \cite{Zeng2019}. In the case of UWB positioning (Pozyx in our trials), indoor obstacles such as reflective surfaces or equipment nearby cause noisy readings. In the case of visual odometers (RealSense T265 in our trials), it is necessary to guarantee that sensor cameras will have rich backgrounds to prevent loss of visual references. The issues of these two technologies may lead to high mutual errors, which are troublesome for Kalman fusion. 

However, since these issues do not co-occur in time, we have designed a novel self-corrective approach that combines the advantages of different technologies when some of them work properly. This approach has three method components: independent Kalman filtering (to avoid the effect of high mutual errors in Kalman fusion), data association by means of stream clustering (to filter out non-gaussian outliers due to intermittent anchor signal loss) and mutual correction of sensor readings based on the generation of cumulative vectors (to avoid the issue of wrong odometer estimations due to lack of visual references). The approach is inspired by the observation that UWB positioning works reasonably well at static spots whereas visual odometer measurements reflect straight displacements correctly even if their lengths are underestimated. 

The self-corrective approach has achieved promising results in our target scenario of quality control imaging for car manufacturing. Even though there is a clear trade-off between positioning accuracy and number of visual odometer restarts (the higher the first the more restarts are necessary), our initial accuracy goal ($\sim$5 cm) was fulfilled. We have also demonstrated that, in this scenario, our approach keeps the benefits of visual odometry by correcting it with UWB data, outperforming the Kalman fusion in \cite{Zeng2019} in case visual odometry fails, both in terms of stopping point accuracy and path RMSE.

As future work we are considering a variant of our approach with a modification of method component A, by combining Kalman fusion of both technologies with independent technology-wise filtering, where the fusion outcome would be used for flight tracking unless the mutual error of independently filtered outputs exceeds a threshold. We will also report experiences in the real industrial scenario in Figure \ref{car}.

\section{ACKNOWLEDGMENTS}
\label{acknowledgment}
The authors are indebted to Mr. Carlos Mesquita, PSA-Citroën Mangualde, and Mr. Alberto Gonz\'alez Casta\~no, Opcon Inform\'atica SL, for sharing their knowledge about the target scenario and validating our laboratory setup.

\newpage

\bibliography{mybibfile.bib}{}
\bibliographystyle{IEEEtran}

\EOD

\end{document}